\documentclass[preprints,article,accept,moreauthors,pdftex]{Definitions/mdpi}
\firstpage{1}
\pubvolume{1}
\issuenum{1}
\articlenumber{0}
\pubyear{XXXX}
\copyrightyear{XXXX}
\datereceived{XX Month Year}
\dateaccepted{XX Month Year}
\datepublished{}
\hreflink{https://doi.org/} 
\pdfoutput=1
\Title{Physical Reservoir Computing Enabled by Solitary Waves and Biologically-Inspired Nonlinear Transformation of Input Data}
\TitleCitation{Physical Reservoir Computing Enabled by Solitary Waves and Biologically-Inspired Nonlinear Transformation of Input Data}

\definecolor{my_green}{rgb}{0.55, 0.71, 0.0}

\usepackage{textcomp}
\usepackage{bm}
\usepackage{comment}

\Author{Ivan S.~Maksymov \orcidA{}}

\AuthorNames{Ivan S.~Maksymov}

\AuthorCitation{Maksymov, I.S.}

\address{%
\quad Artificial Intelligence and Cyber Futures Institute, Charles Sturt University, Bathurst, NSW 2795, Australia; imaksymov@csu.edu.au}

\abstract{Reservoir computing (RC) systems can efficiently forecast chaotic time series using nonlinear dynamical properties of an artificial neural network of random connections. The versatility of RC systems has motivated further research on both hardware counterparts of traditional RC algorithms and more efficient RC-like schemes. Inspired by the nonlinear processes in a living biological brain and using solitary waves excited on the surface of a flowing liquid film, in this paper we experimentally validate a physical RC system that substitutes the effect of randomness for a nonlinear transformation of input data. Carrying out all operations using a microcontroller with a minimal computational power, we demonstrate that the so-designed RC system serves as a technically simple hardware counterpart to the `next-generation' improvement of the traditional RC algorithm.}


\begin{document}
\section{Introduction\label{sec:1}}
A biological brain is a dynamical system characterised by a complex nonlinear and chaotic behaviour at multiple levels \cite{Bab88, McK94, Kor03}. For example, recent theoretical and experimental works have demonstrated that a nerve fibre can operate as a nonlinear waveguide for hybrid electro-acousto-mechanical nerve pulses \cite{Mur00, Hei05, Gon14, Lar14, ElH15, Eng18}. In particular, it has been shown that an interplay between the nonlinearity of the nerve fibres and dispersion processes occurring in them results in the formation and propagation of solitary waves \cite{Hei05, Gon14}.  

It is also well-known that a biological brain can intrinsically process nonlinear acoustic signals such as natural sounds and music \cite{Yu00, Esc02, Lev06, Car19, Kes20}. For example, if a human is exposed to a sound with the spectrum that has all of the acoustic frequency components except the fundamental harmonic, their brain restores the missing frequencies automatically. This phenomenon is called restoration of the missing fundamental \cite{Car96}.

In an experiment involving tests of the auditory system of barn owls \cite{Jan96}, electrodes were introduced into the animal's brain and the owl listened to a version of Strauss’s `The Blue Danube' made up of tones from which the fundamental frequency had been removed. The researchers hypothesised that, if the missing fundamental harmonic was restored at early levels of auditory processing, neurons in the owl’s brain would fire at the rate of the missing fundamental. The experiment confirmed their assumption: the electric output of the electrodes was amplified and played through a loudspeaker, resulting in the original melody of 'The Blue Danube'.

While the biophysical origin of the restoration of the missing fundamental continues to be a subject of debate, it has been suggested that it can be explained by nonlinear and chaotic effects \cite{Chi03}. Nonlinear processes in biological neural systems have also motivated research on artificial neural networks that exploit nonlinear properties of diverse mathematical models and physical systems \cite{Maa02, Ada17, Zho19, Mar20_1, Zha20, Lia20, Has21, He21, Nak22, Lee23, Lop23, Kra23}.   

Inspired by the natural nonlinear processes in a living biological brain and following the recent advances in the field of reservoir computing (RC)---machine learning algorithms for prediction of nonlinear and chaotic time series \cite{Luk09, Nak21, Cuc22}---in this paper we experimentally validate a physical RC system that exploits nonlinear dynamical properties of solitary-like (SL) waves excited on the surface of a flowing liquid film. The resulting neuromorphic computer mimics the functionality of a biological neuron in terms of effectively representing and processing input signals as nonlinear functionals. Yet, it serves as a hardware counterpart to a computationally efficient modification of the traditional RC algorithm called the next generation reservoir computing \cite{Gau21}.

We build and test a technically simple prototype of the proposed physical RC system employing an inexpensive Arduino microcontroller. While the microcontroller has a minimal computational power but the total cost of the prototype does not exceed USD\$100, we argue that in certain practical situations the efficiency of the physical RC system may exceed the one of optimised machine learning software run on a high-performance workstation.
\begin{figure}[t]
 \includegraphics[width=0.9\textwidth]{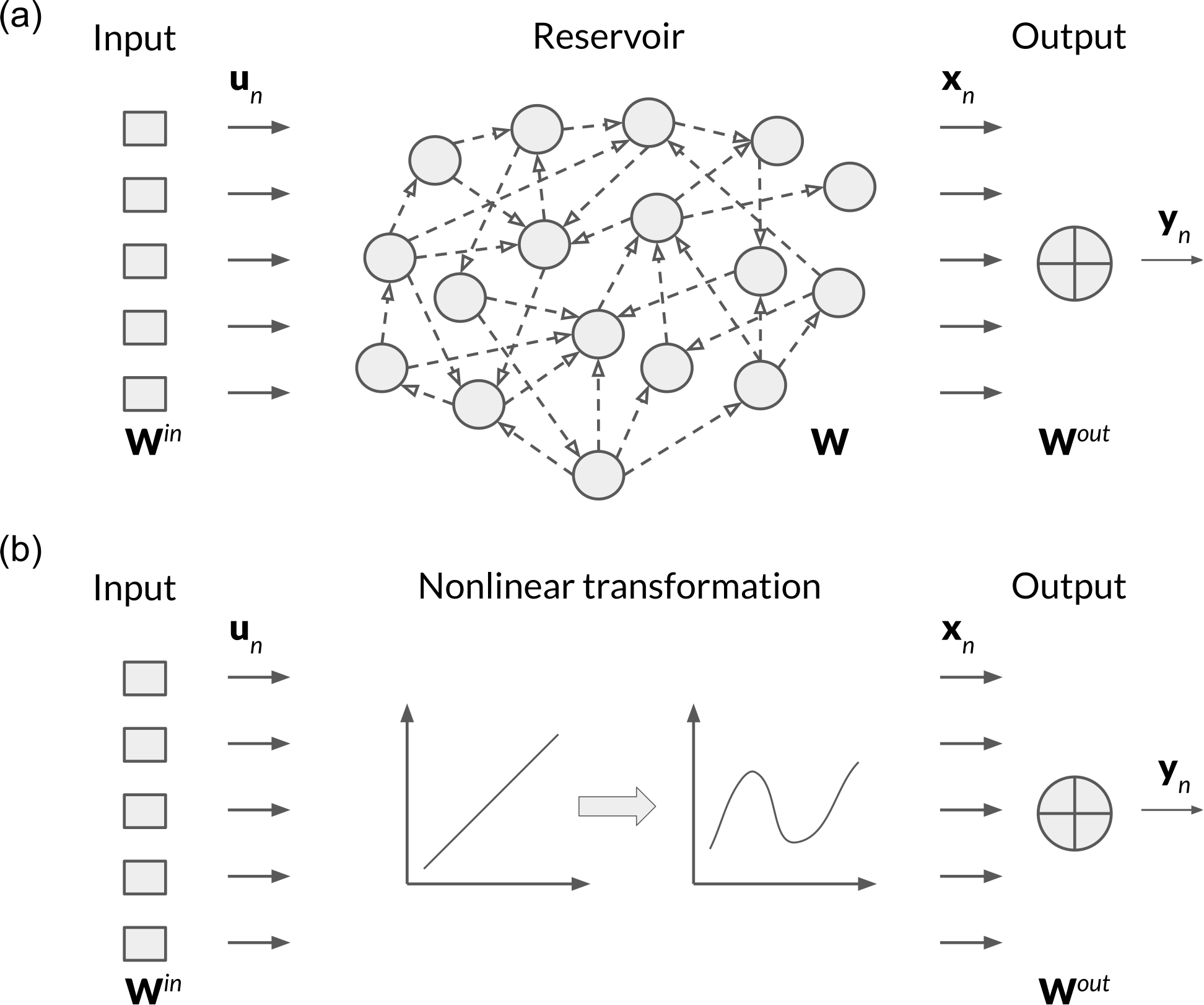}
 \caption{Schematic representation of {\bf(a)} a traditional algorithmic RC system and {\bf(b)} an RC system with a reservoir of random connection substituted by a nonlinear functional of the input data.\label{Fig1}}
\end{figure}

This paper is organised as follows. In Section~\ref{StateArt}, we survey the traditional and next generation RC algorithms as well as overview the recent advances in the field of soliton-based RC systems and adjacent technologies, including analogue and neuromorphic computers based on nonlinear dynamics of fluids. The technical aspects of the experimental setup used in this work are presented in Section~\ref{Setup}. The main results of this paper are presented in Section~\ref{Results} followed by a general discussion in Section~\ref{Discussion}.

\section{State-of-the-art\label{StateArt}}
\subsection{Traditional Reservoir Computing Algorithm\label{TradRC}}
The traditional RC algorithm stems from two independently developed neural network architectures known as Echo State Network (ESN, see Figure~\ref{Fig1}a) \cite{Jae04} and Liquid State Machine (LSM) \cite{Maa02}. A neural network architecture similar to ESN and LSM was also proposed in an earlier work \cite{Kir91} that went mostly unnoticed \cite{Scholarpedia, Mak23_review}.

A typical RC procedure consists of the following computational steps \cite{Luk09}, where $n$ is the index denoting equally-spaced discrete time instances $t_n$:
\begin{enumerate}
  \item Create a vector ${\bf u}_n$ of $N_u$ input values; 
  \item Use a random number generator to define an input matrix ${\bf W}^{in}$ consisting of $N_x \times N_u$ elements and a recurrent weight matrix ${\bf W}$ containing $N_x \times N_x$ elements; 
  \item Calculate the spectral radius and normalise ${\bf W}^{in}$ (see \cite{Luk09, Luk12} for details);
  \item Compute a vector ${\bf x}_n$ of $N_x$ neural activations as
  \begin{eqnarray}
  {\bf x}_{n} = (1-\alpha){\bf x}_{n-1}+
  \alpha\tanh({\bf W}^{in}{\bf u}_{n}+{\bf W}{\bf x}_{n-1})\,,
  \label{eq:RC1}
  \end{eqnarray}
  where hyperparameter $\alpha \in (0, 1]$ controls the update speed of the temporal dynamics;
  \item Construct the state matrix ${\bf X}$ using the values of ${\bf x}_n$;
  \item Train the output as ${\bf W}^{out} = {\bf Y}^{target} {\bf X^\top} ({\bf X}{\bf X^\top} + \beta {\bf I})^{-1}$, where ${\bf I}$ is the identity matrix, $\beta$ is a regularisation coefficient, ${\bf X^\top}$ is the transpose of ${\bf X}$ and ${\bf Y}^{target}$ is matrix composed of target outputs ${\bf y}_n^{target}$ for each time instant $t_n$;
  \item Solve Eq.~(\ref{eq:RC1}) with a new set of target data ${\bf u}_n$ and compute the output vector as ${\bf y}_n={\bf W}^{out}[1;{\bf u}_n;{\bf x}_n]$.
\end{enumerate}

The computations in Step~7 can be organised differently depending on the particular problem under consideration \cite{Luk12}. In the predictive regime (also known as the one-step ahead prediction), Eq.~(\ref{eq:RC1}) is solved for the target data that were not previously seen by the RC system. However, in the generative regime (also known as the free-running forecast), Eq.~(\ref{eq:RC1}) is solved for an updated target data set given by the output generated by the same RC system at the previous time step (i.e.~${\bf x}_{n}$ is calculated using Eq.~(\ref{eq:RC1}) with ${\bf u}_{n} = {\bf y}_{n-1}$).

It is noteworthy that the demonstration of correct operation in the generative mode is a more challenging task compared with the operation in the predictive mode and other regimes where the RC system has access to the expected target data. The same applies to various classification tasks and computational tasks that involve target data that correspond to time-delayed training datasets \cite{Dud23}. On the contrary, in the generative regime the RC system does not know the expected outcome (however, these data can be known to a human operator who is tasked to evaluate the accuracy of predictions made by the RC system \cite{Luk12, Mak21_ESN}). In this context, in the case of the traditional RC algorithm, the operation in the generative mode can be compromised by a spurious data point produced by the RC system \cite{Luk12}. While a number of techniques aimed to improve the generative mode performance of algorithmic RC system have been proposed in the literature \cite{Luk12, Luk21}, the ability of physical RC systems to operate in the generative mode has been difficult to achieve \cite{Raf20, Mak21_ESN, Mak23_EPL}. In the following, we will demonstrate a possibility to create an efficient physical RC system operating in the generative mode.          

\subsection{Next Generation Algorithmic Reservoir Computing}
In the paper \cite{Gau21}, an alternative RC algorithm was proposed where the state matrix ${\bf X}$ of the traditional RC system was substituted by a matrix ${\bf X}^{future}$ consisting of future states ${\bf x}_n^{future}$ corresponding to the current and time-delayed discrete input data points ${\bf u}_n$ and their nonlinear functionals (for a schematic depiction of such an approach see Figure~\ref{Fig1}b). The so-constructed state matrix was then used to calculate ${\bf W}^{out}$, thereby avoiding the computationally demanding Steps 2--4 of the calculation procedure outlined in the previous section. Importantly, based on the results presented in Ref.~\cite{Bol21}, it was also demonstrated that the resulting computational scheme does not only circumvent using matrices of randomly-generated neural connections but is equivalent to and even more computationally efficient than the traditional RC algorithm. This approach was called the next generation reservoir computing \cite{Gau21, Bar22, Kua22, Liu23, Zha23}. Similar algorithms that further improve the performance of RC systems have also been proposed \cite{Ma23}.

The nonlinear part of the future states ${\bf x}_n^{future}$ can be an arbitrary nonlinear function of the input signal. In Ref.~\cite{Gau21}, accurate forecasts were made using  polynomials. It was also established that it suffices to retain just a few polynomial orders to obtain accurate results. A similar result was obtained in Ref.~\cite{Gov22_1} in the context of quantum reservoir computing, where the authors implemented a nonlinear transformation on the input data, demonstrating that further data processing as per the traditional RC algorithm was redundant.

It is noteworthy that a nonlinear functional of the input data can also be obtained using a physical dynamical system. For example, in Ref.~\cite{Mak21_ESN} a chaotic times series was used as a signal that drives nonlinear oscillations of gas bubbles trapped in a liquid. Sinusoidal waves are known to be the fundamental excitations that define the dynamics of many nonlinear systems \cite{Kos93}. In the case of a cluster of bubbles in water, the excitation with a purely sinusoidal wave results in the generation of signals that contain higher order harmonics of the fundamental frequency. This property was used to create a physical analogue of a traditional RC system that approximates a chaotic time series using a large number of sinusoidal waves with different frequencies and amplitudes. 

\subsection{Physical Reservoir Computing Systems Based on Solitary Waves}
Since the traditional RC approach employs the nonlinear dynamics of the mathematical system Eq.~(\ref{eq:RC1}), it was demonstrated that a practicable reservoir could be constructed using a hardware dynamical system \cite{Luk09, Nak21}. Known as the physical RC technique, this approach to computations has been successfully validated, both theoretically and experimentally, resulting in the physical RC systems based on spintronic devices \cite{Rio19, Wat20}, quantum-mechanical systems \cite{Gov21, Dud23, Got23}, electronic circuits \cite{Cow05, Nak21}, photonic systems \cite{Nak21, Sor20, Raf20}, mechanical devices \cite{Cou17} and liquids \cite{Fer03, Gao22, Mak23_review, Mar23}.

It is also well-known that the architectures of some analogue computers used in the 20th~century exploited nonlinear dynamical properties of liquids \cite{Ada19, Sha22, Mak23_review} (e.g.~Ishiguro computer was based on the fundamental nonlinear physical processes that govern the dynamics of solitary and tsunami waves). Consequently, both due to the prior knowledge and independently, there have been proposals of physical RC systems based on the dynamics of solitary waves \cite{Sor19, Jun19, Zen20, Mar20, Sil21, Sha22, Mak23_EPL, Mar23, Mak23_review}.  

Solitary waves propagate with a constant velocity and self-maintain their shape due to an interplay between dispersive processes and nonlinear effects in the medium where they exist \cite{Rem94}. Solitary-like (SL) surface waves that originate from spatio-temporal evolution of flowing liquid films are a particular class of solitary waves known in the field of fluid dynamics (for a review see, e.g., \cite{Mak22, Pot23}). Whereas SL waves are similar to other kinds of solitary waves, they possess a number of unique futures \cite{Liu94}. For instance, unlike two Korteweg-de Vries (KdV) solitary waves that can pass through each other without significant change \cite{Kor95}, two SL waves can merge leading to a more complex behaviour \cite{Liu94}.

The dynamics of both KdV \cite{Mar23} and SL \cite{Mak23_EPL} waves have been shown to be valuable for creating a computational reservoir. In the following, we demonstrate that the specific nonlinear properties of SL waves offer an additional degree of freedom to control the dynamics of the reservoir and help reduce the computational effort while generally reproducing the operation of the traditional RC algorithm.
\begin{figure}[t]
 \includegraphics[width=0.9\textwidth]{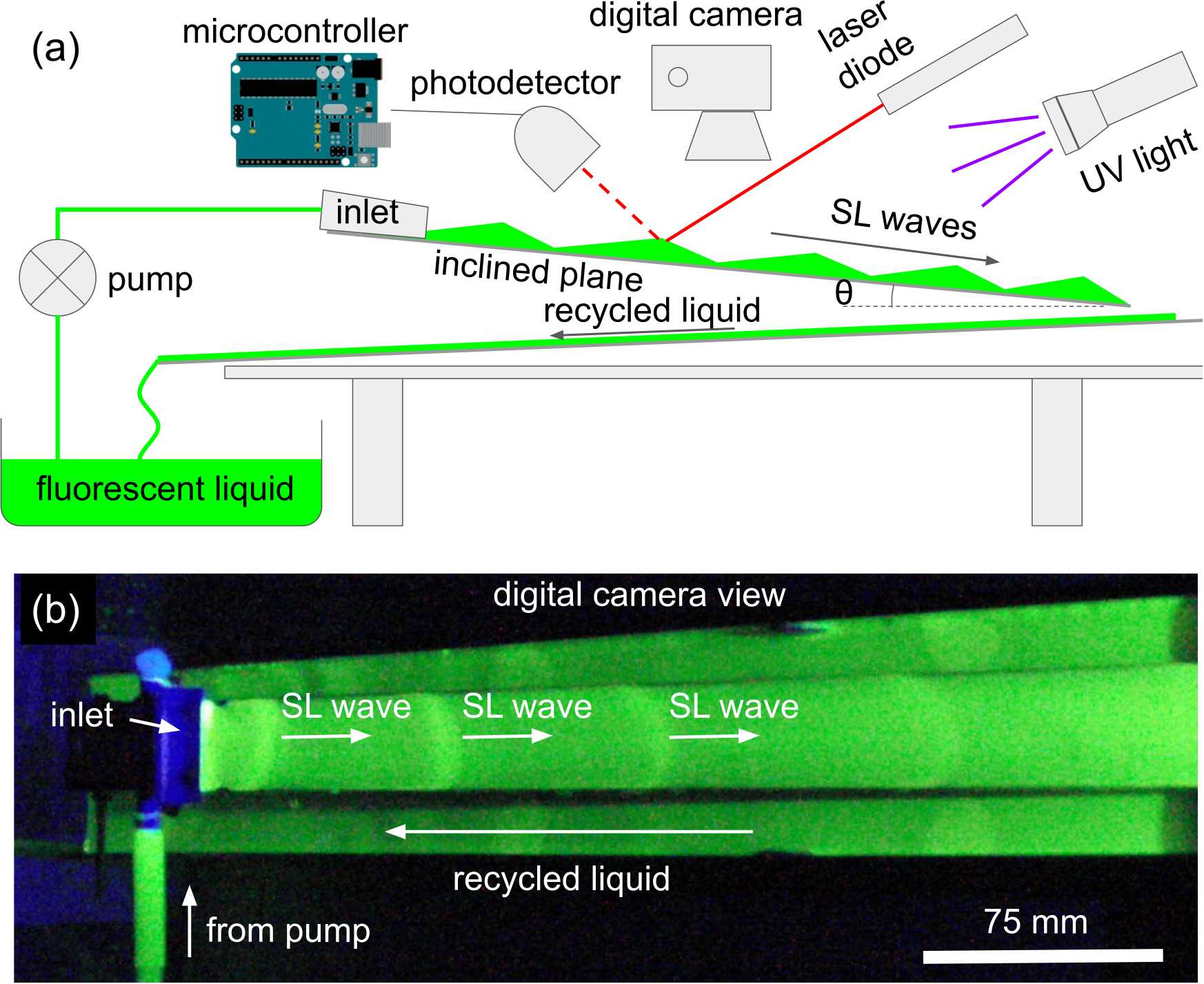}
 \caption{{\bf(a)}~Sketch and {\bf(b)} top view fluorescence photograph of the experimental setup used to validate the proposed architecture of the physical RC system. The fluorescent dye, UV light and digital camera play an auxiliary role and can be removed from the setup without compromising its operation. The remaining components of the setup are controlled by an Arduino microcontroller that is also used to process the raw data traces.\label{Fig2}}
\end{figure}

\section{Experimental Setup\label{Setup}}
Figure~\ref{Fig2} shows the experimental setup of the proposed RC system that uses tap water as the operating liquid. The experiment was designed so that the setup enabled both studies of a prototype of the RC system and investigations of the fundamental physical properties of SL waves. In the studies of the RC system, the SL waves were detected using a customised red laser diode-photodetector pair. To better understand the physical properties of SL waves, an organic fluorescent dye was dissolved in water and the liquid was illuminated with UV light. A high-speed digital camera was used to record the fluorescence images (Figure~\ref{Fig2}b). The addition of the fluorescent dye did not change the fluid-mechanical properties of the liquid nor it noticeably the operation of the photodetector.

The SL waves were excited on the surface of a liquid film flowing along an elongated metal plate inclined with respect to the ground by the angle $\theta=3$\,$^{o}$ \cite{Mak22}. To control the SL waves, the flow rate of the liquid was varied using a miniature electric pump. The electric signal that powered the pump corresponded to the input values ${\bf u}_n$ in the algorithmic RC system. To match the temporal dynamics of the input signal with the temporal dynamics of the SL waves, that signal was downsampled so that the fundamental frequency in its spectrum was 1--2\,Hz depending on the particular experimental scenario. The so-controlled pump enabled a smooth variation in the amplitude of the created SL waves.   

All components of the setup were controlled by an Arduino~UNO~R3 microcontroller (16~MHz clock speed, 2~kB~RAM) running customised software. The same microcontroller was used to process raw experimental data traces and produce the final results. A personal computer was used only to produce the figures presented in this paper.
\begin{figure}[t]
 \includegraphics[width=0.99\textwidth]{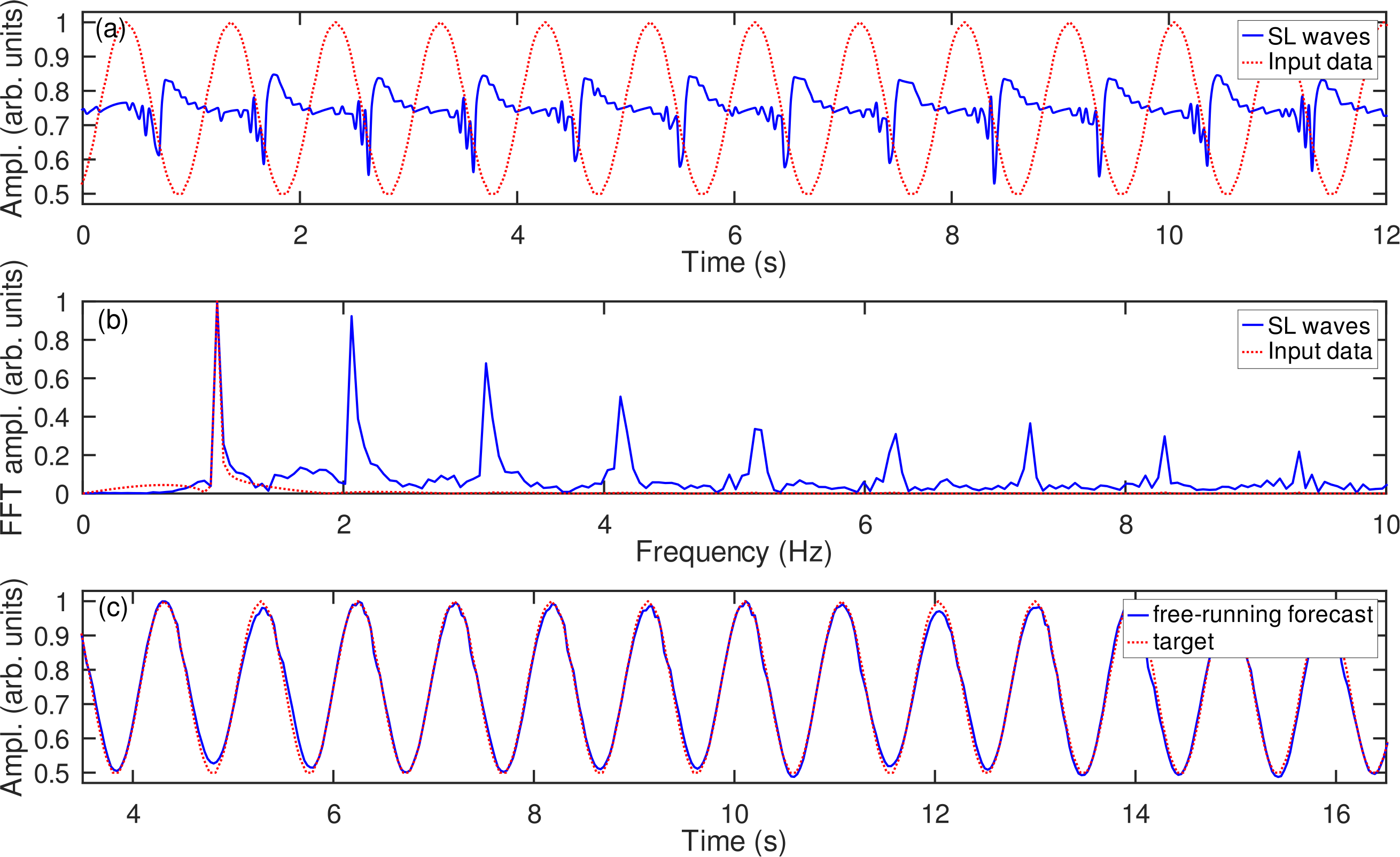}
 \caption{{\bf(a)}~Input sinusoidal signal (the dotted curve) and the SL waves excited by it (the solid curve). {\bf(b)}~Fourier spectra of the signals in Panel~(a). {\bf(c)}~Free-running forecast of the future evolution of the sinusoidal waves made by the RC system based on the SL waves. Note that the timescale in Panel~(c) is unrelated to that in Panel~(a).\label{Fig3}}
\end{figure}

\section{Results\label{Results}}
\subsection{Formation of the Nonlinear Functional}
The nonlinear transformation of the input data implemented in the RC system proposed in this work is illustrated in Figure~\ref{Fig1}b. To experimentally validate this approach, we choose a simple but non-trivial test problem \cite{Tro20} of prediction of the future evolution of a sinusoidal wave. In the framework of our experiment, this scenario corresponds to the excitation of SL waves using the pump that is driven by a 1\,Hz sinusoidal electric signal.

As shown in Figure~\ref{Fig3}a, each period of the sinusoidal wave triggers the generation of an SL wave (also see Supplementary Video~1). Each SL wave consists of a main pulse with a steep front that is preceded by a train of secondary pulses of a smaller amplitude \cite{Liu94, Mak22}. When an SL wave moves downstream, the photodetector first receives the light reflected from the secondary pulses and then it senses the main pulse. As a result, in Figure~\ref{Fig3}a we observe a reversed picture: the train of low-amplitude secondary pulses precedes the main pulse.

We compute the Fourier spectra of both sinusoidal and SL wave signals and plot them in Figure~\ref{Fig3}b. We can see that the spectrum of the SL waves is composed of the fundamental (1\,Hz) frequency peak and the peaks that correspond to the second, third, fourth and so on higher-order harmonics. The spectrum of the sinusoidal wave has only one frequency peak at 1\,Hz. Thus, we can see that the SL waves effectively represent the input signal as a nonlinear polynomial function. 

Empirically, the nonlinear generation of the higher-order harmonic frequency peaks in Figure~\ref{Fig3}b can be explained drawing an analogy between the steep front of the main pulse of the SL waves and large-amplitude shock-like acoustic disturbances \cite{Bal04, Mak22}. Shock waves are known to have a frequency spectrum composed of a large number of higher-order harmonic frequencies \cite{Mak19}. A mathematical insight into the nonlinear transformation process can also be gained analysing nonlinear partial differential equations that approximate complex nonlinear physical phenomena \cite{Kur22}. 

We use the SL wave signal as the vector ${\bf x}_n$ of neural activations of the traditional RC algorithm (Figure~\ref{Fig1}a). Importantly, although in other physical RC systems the output of the physical reservoir is also often interpreted as a vector ${\bf x}_n$ (see, e.g., \cite{Wat20, Mak23_review}), in our SL-wave based RC system the vector ${\bf x}_n$ contains the values that are both time-delayed and a nonlinear function of the input data. Subsequently, this vector is conceptually similar to a future vector used in the next generation RC algorithm \cite{Gau21}. Moreover, while only the quadratic nonlinear term was retained in Ref.~\cite{Gau21} to reduce the computational effort, the use of SL waves enables us to retain many nonlinear terms as shown in Figure~\ref{Fig3}b.   

Yet, unlike the original next generation RC algorithm \cite{Gau21} and it modifications, in our RC system the vector ${\bf x}_n$ is constructed only once as part of the training stage and it is not updated at the exploitation stage. That is, our algorithm does not follow Step~7 in the traditional RC scheme where Eq.~(\ref{eq:RC1}) needs to be solved. This simplification aligns with the demonstrations of the primary role of the nonlinear transformation of the input data in Ref.~\cite{Gov22_1, Mak21_ESN} and it is discussed in more detail in the next subsection.

Figure~\ref{Fig3}c shows the results of the free-running forecast (generative mode) made by the SL wave physical RC system using the procedure described above. We can see that the RC system correctly predicts the future evolution of the sinusoidal wave. We also note that this result was obtained with a minimum post-processing, involving only the removal of the DC component of the output signal.
\begin{figure}[t]
 \includegraphics[width=0.99\textwidth]{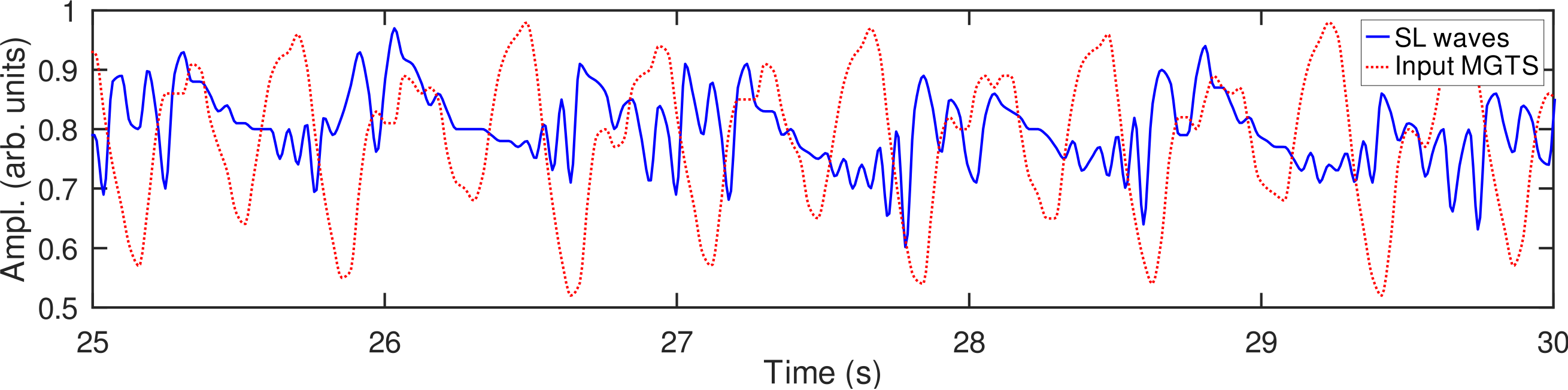}
 \caption{Input MGTS signal (the dotted curve) and the SL waves excited by it (the solid curve). Unlike in Figure~\ref{Fig3}a, since each variation of MGTS results in the generation of SL waves with different amplitude and propagation speed, the SL waves collide and form more complex wave profiles.\label{Fig3_1}}
\end{figure}

\subsection{Advantages for Generative Mode Operation}
Before we test the physical RC system on a more challenging task of chaotic time series prediction, we discuss one particular advantage of the neuromorphic computation approach introduced in the previous section---making a forecast using neural activations ${\bf x}_n$ constructed at the training stage. 

In the generative regime, the implementation of Step~7 of the computational procedure outlined in Section~\ref{TradRC} corresponds to the introduction of a feedback loop from the output of the RC system to its input. Subsequently, the reservoir can be considered to be a self-oscillator \cite{Jen13}, which is an established fact \cite{Sho23}. 

Self-oscillations have been obtained in diverse physical system and mathematical models that involve an oscillator that uses its own output signal to modulate the phase of the external driving force. The so-constructed system can maintain a periodic motion using a source of power that lacks periodicity \cite{Jen13}.

We demonstrate that the introduction of a feedback can be avoided without compromising the operation of the RC system. This approach is based on the following facts. A key component of the operation of a trained reservoir is the matrix ${\bf W}^{out}$ that remains unchanged at the exploitation stage and is used to make a forecast as ${\bf y}_n={\bf W}^{out}[1;{\bf u}_n;{\bf x}_n]$. From the physical point of view, the chief role of the feedback loop is the maintenance of an appropriate temporal dynamics of the reservoir.

We established that an RC system can produce plausible results when the reservoir is driven by a signal that is similar to the training dataset in terms of temporal dynamics and magnitude. We successfully tested signals corresponding to a delayed version of the training and synthetic signals corresponding to a sum of sinusoidal waves with different frequencies and amplitudes (for a relevant discussion see \cite{Mak21_ESN}). Thus, provided that the dynamics of the reservoir is maintained with an appropriate rate and amplitude, by virtue of the values of ${\bf W}^{out}$ the reservoir can produce feasible results.  

This finding is especially useful in the case of physical RC systems. Indeed, the introduction of a feedback loop in a computer code that implements the traditional RC algorithm does not present significant technical difficulties \cite{Luk12}. However, in physical RC systems where the dynamics of the reservoir is controlled by an electronic, optical or opto-electronic circuit \cite{Sor20, Raf20}, apart from certain technical limitations a feedback loop introduces a time delay \cite{Raf20}. Such a delay can be longer than the timescale of the reservoir dynamics and it can interrupt the dynamics of the reservoir, requiring the application of complex experimental techniques aimed to restore the intended reservoir dynamics \cite{Raf20}. (We also confirmed that the introduction of an artificial delay in the traditional RC algorithm compromises the operation of the reservoir.)

Hence, the technical simplification proposed in this paper enables the developers of physical RC systems to avoid the use of feedback loops, also simplifying the design of the device and decreasing its cost. While any simplification comes at a cost, in the following we demonstrate that our physical RC system can undertake complex tasks, producing practicable forecasts.     

\subsection{Free-running Forecast of Chaotic Time Series}
As a next step, we demonstrate the ability of the physical RC system based on SL waves to predict a Mackey-Glass time series (MGTS), which is a standard test problem used to verify the accuracy of neural network models \cite{Rod10, Luk12, Mor21}. While other chaotic time series \cite{Lor63, Ros76, Hen76, Ike79, Gau21} have been employed to test both algorithmic and physical RC system, in our previous work \cite{Mak21_ESN} we demonstrated that, due to the complexity of the Mackey-Glass model, it suffices to test an RC system using MGTS to reasonably expect that the same RC system will be able to process other time series with acceptable accuracy.          

We generate an MGTS dataset solving the delay differential equation \cite{Mac77}
\begin{eqnarray}
  \dot{x}_{_{MG}}(t)
  &=&\beta_{_{MG}}\frac{x_{_{MG}}(\tau_{_{MG}}-t)}
      {1+x_{_{MG}}^{q}(\tau_{_{MG}}-t)}-\gamma_{_{MG}}x_{_{MG}}(t)\,,
  \label{eq:MG}
\end{eqnarray}
where overdot denotes differentiation with respect to time and $\tau_{_{MG}}=17$, $q=10$, $\beta_{_{MG}}=0.2$ and $\gamma_{_{MG}}=0.1$ \cite{Luk12}. Then we split the resulting dataset into two parts. The first part that corresponds to a few (typically 5--6) pseudo-periods of variation of MGTS is used to train the RC system. The longer second part is used as the target data that are not known to the RC system but used exclusively to evaluate the accuracy of the forecast made by the RC system in the generative regime. We underscore that this approach differs from the standard one where the training and target datasets have the same length, showing that shorter training datasets can be used to train the RC system based on SL waves.
\begin{figure}[t]
 \includegraphics[width=0.99\textwidth]{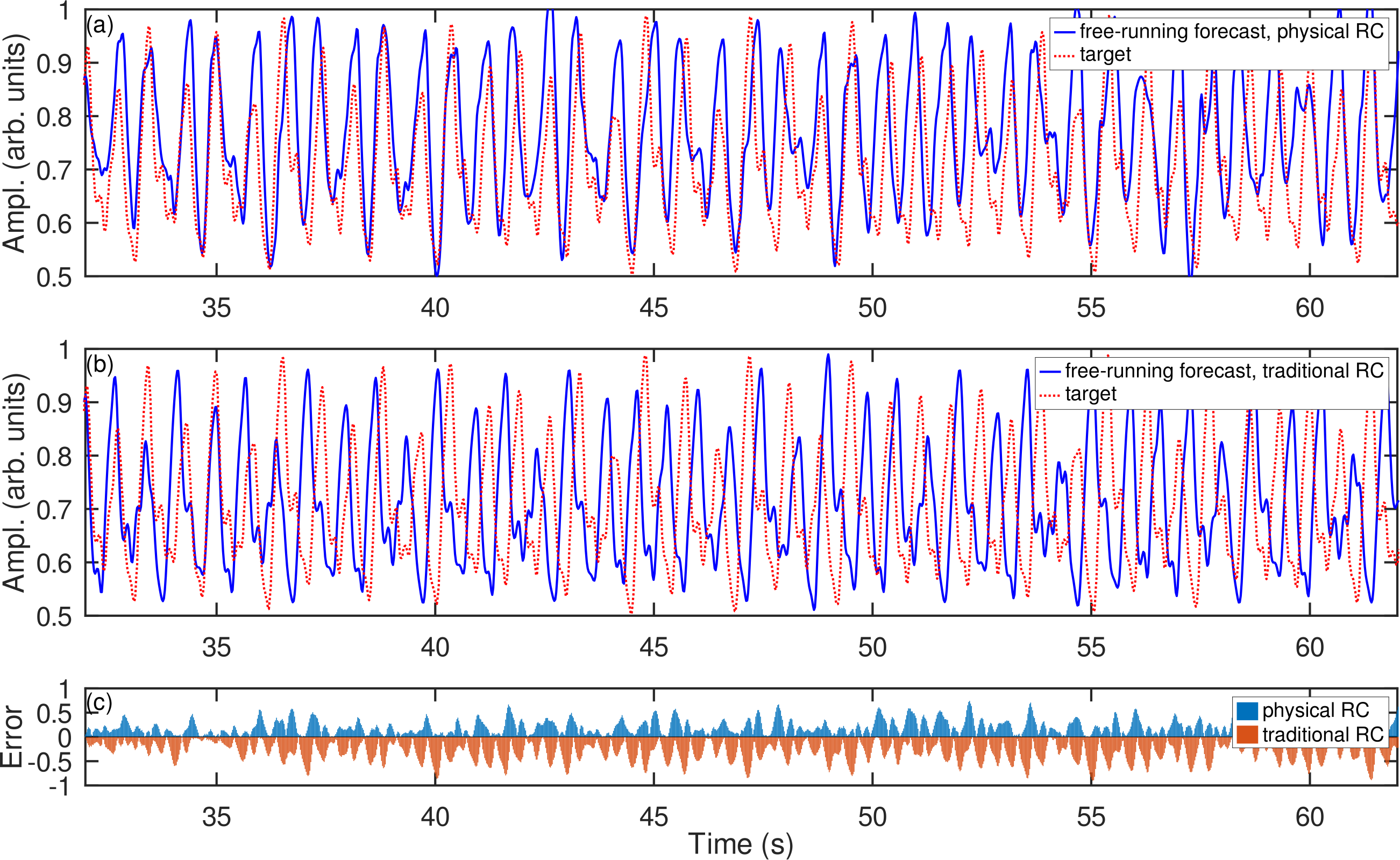}
 \caption{Generative mode operation (free-running forecast) of {\bf(a)}~physical RC system based on SL waves and {\bf(b)}~traditional algorithmic RC system (the solid curve) compared with the target MGTS (the dotted curve). {\bf(c)}~Modulus of absolute error of the forecasts produced by the physical and traditional RC algorithmic systems. Note that for the sake of comparison the error of the traditional RC system is plotted with the negative sign.\label{Fig4}}
\end{figure}

Unlike the SL waves produced by each period of the sinusoidal signal in Figure~\ref{Fig3}a, the SL waves produced by each oscillation in the MGTS signal have different shapes and amplitudes (Figure~\ref{Fig3_1}; for a theoretical analysis see Ref.~\cite{Mak23_EPL}). As demonstrated in Refs.~\cite{Mak22, Pot23}, the SL waves of different amplitude have different propagation speeds. Previously we demonstrated that this physical property enables the SL waves to interact one with another, resulting in a complex nonlinear dynamical behaviour that is suitable for creating a computational reservoir with a short-term memory \cite{Mak23_EPL}.  

The free-running forecast made by the physical RC system is presented in Figure~\ref{Fig4}a. In Figure~\ref{Fig4}b we also plot the free-running forecast made by the traditional algorithmic RC system implemented following the algorithm presented in Ref.~\cite{Luk12}. Both systems were trained and tested on the same training and test datasets, respectively.

It is noteworthy that the traditional RC system requires a much longer training dataset compared with the one needed for the physical RC system. Therefore, in the particular example shown in Figure~\ref{Fig4}a,~b, the length of the training dataset was increased to enable the traditional RC system to make a meaningful forecast (i.e.,~unlike the algorithmic RC system, the physical RC system can operate using shorter training datasets, which is a property that was previously observed mostly in quantum RC systems \cite{Dud23}). We also established that, firstly, the traditional RC system requires a reservoir with at least 1000\,neurons to produce an interpretable result and, secondly, any further increase in the number of neurons leads to overtraining of the reservoir (the so-called overfitting condition \cite{Luk09}), rendering the RC system unable to reproduce the dynamics of the target time series. 

Thus, we can see that the forecast made by the physical RC system reasonably captures the long-term variation of MGTS in general. Yet, overall, the accuracy of the forecast made by the physical RC is higher than that of the traditional RC algorithm (Figure~\ref{Fig4}c). We emphasise that in this test the traditional RC system used 1000\,neurons and its operation required approximately 2\,seconds of CPU time of a high-performance workstation computer (Mac Studio M1, Ultra 20-core CPU, 128\,GB RAM).

This analysis made above reflects the well-known fact that, after a certain threshold, any further increase in the size of a computational reservoir results in a small, if any, increase in the performance \cite{Cuc22}. The physical RC system presented in this work does suffer from this drawback since it does not rely on random matrices used in the traditional RC algorithm.

The problem of saturation of the reservoir size also does not exist in the framework of the next generation RC algorithm \cite{Gau21}. However, software that implements it also requires a modern computer. Yet, the computational effort associated with the next generation RC computations increases as the nonlinear functional of the input data becomes a higher-order polynomial compared with a quadratic functional used in Ref.~\cite{Gau21}. On the contrary, the physical RC system based on SL waves automatically represents the input data using many higher-order nonlinear harmonics and, therefore, does not require any further adjustment of the input datasets.

\section{Discussion\label{Discussion}}
\subsection{Energy Efficiency, Power Consumption and Cost}
Thus, while a high-accuracy long-term forecast of MGTS using a traditional RC algorithm requires a high-performance workstation, the result in Figure~\ref{Fig4}b was obtained using a rather modest auxiliary computational power of an Arduino microcontroller (Arduino microcontrollers used in the previous demonstration of physical RC systems \cite{Kan21} were not employed to post-process data). Indeed, employing floating point operations per second (FLOPS) as a unit of measure, we estimate the maximum performance of the Arduino microcontroller used in this work to be of order of 0.1\,MFLOPS. For comparison, the workstation computer used to test the traditional RC system can readily deliver more than 2\,TFLOPS. Furthermore, the Arduino microcontroller has just 2\,kB~RAM compared with a 128~GB~RAM of the workstation and the entire experimental setup shown in Figure~\ref{Fig2} consumes less than 1\,W of power compared with a more than 200\,W power consumption by the workstation. 

While, in principle, less powerful workstation models can be used instead of the workstation used in this paper, the cost of a computer with a minimal specification needed to run the software that implements the traditional RC algorithm is around USD\$1,000 but the power consumed by it is about 50\,W. On the contrary, the cost of the prototype of the physical RC system is just USD\$100.

To put the cost and power consumption of the physical RC system further into perspective, we note that the price of a mass-produced Akida$^{\rm{TM}}$ PCIe Board with a BrainChip neuromorphic processor is USD\$499 but an assembled `Development Kit' system based on a personal computer costs USD\$9,995 and consumes 180\,W \cite{brainchip}. Generally speaking, in light of an exponential increase in the computing power demand seen in the last decade \cite{Meh22}, the liquid-based unconventional computing system appears to be a plausible alternative to conventional microelectronics \cite{Khe22}. Unconventional liquid-state computational systems can also outperform emergent photonics-based computers \cite{Sha20} in terms of energy efficiency since the latter may require high-intensity laser light to induce nonlinear effects needed for a physical implementation of a neural network \cite{Sor19} but nonlinear processes in liquids can be obtained virtually effortlessly \cite{Mak19}.   

\subsection{Potential Applications}
Admittedly, the accuracy of the free-running forecast made by the physical RC system may not be suitable for quantitative analysis such as mathematical modelling of financial markets. Nevertheless, a number of works demonstrated a high value of qualitative forecasts of variation of financial markets \cite{Lin09, Sun21_market}, which is a finding that aligns with the scope of financial physics, an academic discipline that studies financial markets as physical systems, thus complementing quantitative finance by elucidating the physical nature of financial nonlinear dynamical processes \cite{Lil00}.

However, the main advantage of the proposed physical RC system does not necessarily come from its comparison with the computational algorithms used in quantitative research. In fact, the concept of reservoir computing has been applied to study neural information processing in biological brain networks \cite{Maa02, Kaw19, Sua21, Dam22, Rao22} where the requirement for an RC system to make quantitatively exact forecasts can be relaxed. On the other hand, it is mandatory that the RC system operates similarly to a biological brain. Importantly, this requirement does not only mean a functional resemblance to a brain but also implies a brain-like energy efficiency \cite{Rao22}. Yet, since it has been suggested that a healthy biological brain relies on non-randomness of neural connections (i.e.~randomness may be associated with some disorders of the nervous system) \cite{Spo11}, an RC system that does not use random matrices should serve as a better model of biological neural networks than a traditional RC algorithm.

In this context, the physical RC system based on SL waves has certain advantages. Indeed, firstly, the physical RC system does not rely on randomness. Secondly, as discussed in the Introduction, the nonlinearity of SL waves employed in the RC system is physically similar to the nonlinear processes in the nervous system. Thirdly, the physical system is both computationally and energy efficient. Fourthly, it is conceivable that SL waves or other types of solitary waves \cite{Gon14} could be used to create an artificial neuron. The research work in this direction is going on both from the biophysical and chemical \cite{Khe22} and machine learning \cite{Kaw19, Sua21, Dam22} points of view.

It is also noteworthy that biological environments represent significant technological challenges for the developers of implantable brain-computer interfaces and other AI-based system intended co-operate with the nervous system of a living organism \cite{Sah21}. These challenges include the impact of various physical processes such as vibrations, scattering and absorption. Previously, we established that SL waves are highly immune to external mechanical vibrations at the frequencies from 20 to 100\,Hz \cite{Mak22}. Therefore, the RC system based on SL waves can also operate in noisy environments. Yet, its ability to make forecast should not be affected by strong magnetic and electromagnetic fields, which is, for example, a technological challenge for spin-wave-based neuromorphic computers \cite{Wat20, Pap21, Nak23}.       

Finally, the fields of creative music composition and manipulation of sound \cite{Mak23_music} can also benefit from the experimental techniques employed in proposed physical RC system. For example, one can use the experimental setup involving SL waves excited on the surface of a flowing liquid to create artistic music effects \cite{Lau07}. This intriguing application returns us to the discussion of nonlinear effects associated with our perception of sound and music (see the Introduction), thereby demonstrating a fundamental link between nonlinear dynamics, natural intelligence and AI. 

\section{Conclusions}
We have demonstrated an experimental physical RC system that employs solitary waves to implement a biologically-inspired nonlinear transformation of input data instead of large matrices of random neural connections that are central to the traditional RC algorithm. Post-processing raw experimental data using a technically simple and inexpensive Arduino microcontroller, we built a practicable neuromorphic computer that costs less than USD\$100 and consumes very small electric power compared with both standard digital computers and many commercial and experimental neuromorphic systems that employ electron and photonic devices to mimic the neurons.

While feasible fluidic microprocessors have been demonstrated \cite{Lee21} thus paving the way for further optimisation, miniaturisation and eventual commercialisation of the physical RC system demonstrated in this paper, we foresee the application of the ideas proposed in this work in neuromorphic systems designed to closely resemble a biological neuron \cite{Sar22}. Indeed, since a living organism contains a significant amount of water and, therefore, can respond to physical stimuli in a nonlinear-dynamical manner \cite{Mak20_worm}, it is plausible that an SL wave based RC system can be implemented as an organic bio-fluid-based artificial neuron that can communicate with biological neurons to complement and enhance the natural functionality.

Last but not least, the SL wave-based neuromorphic computing platform is remarkably technically simple and, at the same time, rich in physical effects. Therefore, it can be both further explored by scientists and used by high-school and undergraduate students to understand the principles of neuromorphic computing \cite{Rou21}.  

\vspace{6pt}

\authorcontributions{I.S.M.~conceived the idea, built the experimental setup, conducted experiments and prepared the manuscript.}

\funding{The author did not receive any funding to support this project.}

\institutionalreview{Not applicable.}

\informedconsent{Not applicable.}

\dataavailability{All data generated in this work are presented in this article.}

\acknowledgments{The author acknowledges useful discussions with Dr Abbas Hussein and Associate Professor Andrey Pototsky.}

\conflictsofinterest{The author declares no conflict of interest.}

\abbreviations{Abbreviations}{
The following abbreviations are used in this manuscript:\\

\noindent 
\begin{tabular}{@{}ll}
 artificial intelligence &AI\\
 direct current &DC\\
 Echo State Network &ESN\\
 floating point operations per second &FLOPS\\
 Korteweg-de Vries &KdV\\ 
 Liquid State Machine &LSM\\
 Mackey-Glass time series &MGTS\\
 reservoir computing &RC\\
 random access memory &RAM\\
 solitary-like &SL\\
 ultraviolet &UV\\ 
 United States Dollar &USD\\ 
  
\end{tabular}
}


\begin{adjustwidth}{-\extralength}{0cm}
\reftitle{References}


\externalbibliography{yes}
\bibliography{refs}

\end{adjustwidth}
\end{document}